\documentclass[akbc,twoside,11pt,lettersize]{article}
\usepackage{akbc}
\usepackage{subcaption}
\usepackage{graphicx}
\usepackage{multirow}
\usepackage{tabularx}
\usepackage{float}
\usepackage{lingmacros}
\usepackage{booktabs}
\usepackage{mathtools}

\usepackage[T1]{fontenc}
\usepackage[utf8]{inputenc}
\usepackage{microtype}
\usepackage{fixltx2e}
\usepackage{wrapfig}

\newcommand{\newcite}[1]{\citeauthor{#1} [\citeyear{#1}]}

\ShortHeadings{Annotating Implicit Reasoning in Arguments with Causal Links}{Singh, Inoue, Mim, Naitoh \& Inui}
\finalcopy % Uncomment for camera-ready version, but NOT for submission.
\begin{document}

\title{Annotating Implicit Reasoning in Arguments\\ with Causal Links}

\author{\name Keshav Singh \email singh.keshav.t4@dc.tohoku.ac.jp \\
        \addr Tohoku University 
        \AND
        \name Naoya Inoue \email naoya.inoue.lab@gmail.com \\
        \addr Stony Brook University / RIKEN
        \AND
        \name Farjana Sultana Mim \email mim.farjana.sultana.t3@dc.tohoku.ac.jp\\
        \addr Tohoku University
        \AND
        \name Shoichi Naitoh \email naito.shoichi.t1@dc.tohoku.ac.jp \\
        \addr Ricoh Company, Ltd. / RIKEN / Tohoku University
        \AND
        \name Kentaro Inui \email inui@tohoku.ac.jp\\
        \addr RIKEN / Tohoku University}

\maketitle

%Intro change from Inui-san --> our study related to argkG workshop, how ? Number of studies in other domains on auto-exxtraction of knowledge, our focus is on CS. Our knowledge is not explicitly statted in arg but implicitly presupposed. (not easy to automatize this therefore as an alternative)
% End of paper (conclusion), mention of arggraph 

\begin{abstract}
Most of the existing work that focus on the identification of implicit knowledge in arguments generally represent implicit knowledge in the form of commonsense or factual knowledge.
However, such knowledge is not sufficient to understand the implicit reasoning link between individual argumentative components (i.e., claim and premise).
In this work, we focus on identifying the implicit knowledge in the form of argumentation knowledge which can help in understanding the reasoning link in arguments.
Being inspired by the \emph{Argument from Consequences scheme}, we propose a semi-structured template to represent such argumentation knowledge that explicates the implicit reasoning in arguments via causality.
We create a novel two-phase annotation process with simplified guidelines and show how to collect and filter high quality implicit reasonings via crowdsourcing.
We find substantial inter-annotator agreement for quality evaluation between experts, but find evidence that casts a few questions on the feasibility of collecting high quality semi-structured implicit reasoning through our crowdsourcing process.
We release our materials (i.e., crowdsourcing guidelines and collected implicit reasonings) to facilitate further research towards the structured representation of argumentation knowledge.
\end{abstract}

\section{Introduction}
\label{Introduction}
In daily conversations, humans create many arguments that are grounded on knowledge which is often left implicit or unstated~\cite{ennis1982identifying}. 
Consider the following example of an argument consisting of two argumentative components: \textit{Claim} (i.e., declarative statement) and its \textit{Premise} (i.e., supporting statement):

{\enumsentence{\textbf{Claim}: We should ban surrogacy.\\
\textbf{Premise}: Surrogacy often creates abusive and coercive conditions for women.} 
\label{ex:1}}

%Challenge
In Example~\ref{ex:1}, one can easily utilize commonsense to infer \emph{Surrogacy is bad for women} or \emph{Surrogacy directly pertains to women}
% \emph{Banning surrogacy would lessen abusive or coercive conditions for women}
which is implicitly asserted in the argument. 
While explication of such implicit knowledge in arguments is an important component for understanding the facts on which the argument is build on~\cite{hulpus2019towards, becker2020explaining}, such commonsense knowledge is not sufficient to help the reader understand how the premise supports the claim, i.e., it is unclear \emph{why/how banning surrogacy would lessen the abusive and coercive conditions for women}.
In order to answer this question, one needs to understand the argumentation knowledge, i.e., the implicit reasoning link between the two argumentative components, which is generally not explicated by commonsense knowledge.
For example, the following explication: \emph{"Banning surrogacy causes decrease in number of women working as surrogates which lessens the abusive and coercive conditions for women"} is necessary for understanding the underlying reasoning between the claim and premise.

\begin{figure}[t]
    \centering
    \includegraphics[width=\linewidth]{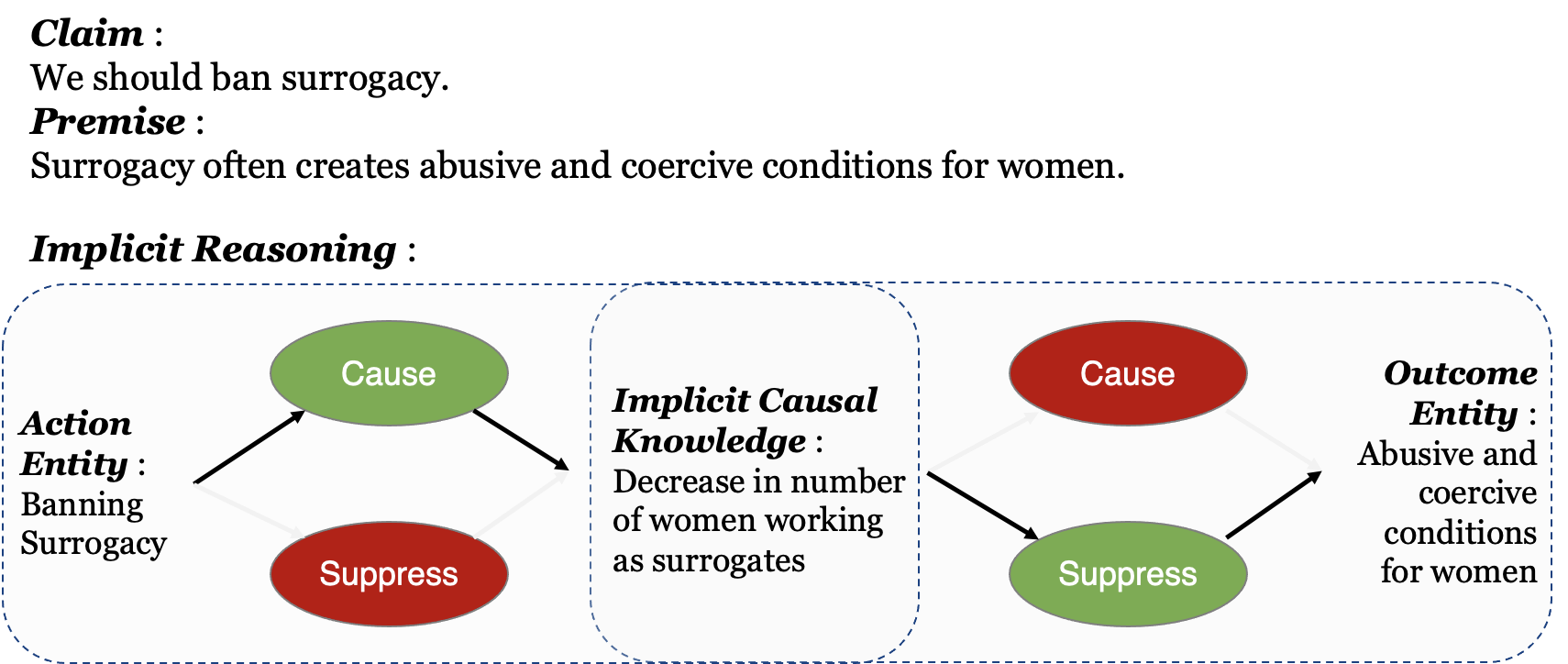}
    \caption{An example of our proposed semi-structured format to explicate implicit reasoning in arguments.}
    \label{fig:intro_eg}
\end{figure}

In this work, we propose an annotation scheme for identifying the implicit knowledge between claim and premise in the form of argumentation knowledge, which explicates their implicit reasoning link.
Specifically, as shown in Figure~\ref{fig:intro_eg}, we create a semi-structured template to represent the implicit reasoning between claim and premise via causality.
We draw our inspiration for creating such template from the \emph{Argument from Consequences Scheme}~\cite{walton2008argumentation} which has been shown to be useful for explicating implicitly asserted propositions~\cite{feng-hirst-2011-classifying,reisert-etal-2018-feasible,al2020end} in arguments.
Our key intuition behind this form of representation is to capture and explicate the relevant logic flow between claim and premise via causality so that we can understand why/how the outcome in premise \texttt{("abusive \_\_\_ for women")} is justified by carrying out the action in claim \texttt{"(banning surrogacy)"}.
% However, for identifying the implicit reasoning link, such schemes need to be carefully utilized with simplified guidelines to ensure correct annotation of implicit reasoning such that it explicates the logical connection between claim and premise. 

Numerous prior works have demonstrated the explication of implicit knowledge in arguments either in the form of commonsense knowledge~\cite{becker2017enriching} or as argumentation knowledge to capture the implicit reasoning link between claim and premise~\cite{boltuzic-snajder-2016-fill,habernal-etal-2018-semeval}. 
However, the implicit reasonings are usually explicated in unstructured form (i.e., free-text or natural language sentence), having no restriction on their lexical structure which sometimes makes it difficult to interpret how the implicit reasonings relate to any of the information given in the claim and premise.
In contrast to that line of work, our proposed implicit reasonings is semi-structured and our work is inspired from Argument from Consequences scheme where we focus on causality to explicitly relate the implicit reasoning with key information given in claim and premise.

The most similar research to our work was done by~\newcite{saha2021explagraphs}, who attempt to create explanation graphs to reveal the reasoning process involved in explaining why a premise supports its claim by explicating very fine-grained implicit commonsense knowledge in the argument.
In contrast, our work differs in multiple ways and makes the following contributions: (1) Our work additionally covers \emph{suppress} causal relation with focus on argumentation schemes, (2) our annotation considers multiple possible ways to explicate implicit reasoning in arguments with diverse implicit knowledge while \newcite{saha2021explagraphs} focus on a single explanation graph per argument and (3) we provide an in-depth analysis of quality and coverage of annotated implicit reasonings.

% In contrast, we additionally perform quality annotation of implicit reasoning on a scale of 0-5 with specific guidelines and report the results to depict the overall quality of our annotations. 
% To this end, we create a novel two phase annotation process with simplified guidelines to (i) crowdsource annotations of implicit reasonings on a large scale and (ii) filter high quality implicit reasonings through crowd based qualitative evaluation.

\section{Semi-structured Implicit Reasoning}
\label{sec-3}
In contrast to explicating implicit knowledge in arguments with general facts or commonsense in unstructured form, we are interested in framing implicit knowledge in the form of argumentation knowledge that is specifically needed to understand the underlying reasoning link between claim and premise.
In particular, 
% inspired from Argument from Consequences scheme~\cite{walton2008argumentation},  
as shown in Figure~\ref{fig:intro_eg}, we construct a template for explicating such implicit reasonings with causality (i.e., \emph{cause/suppress}) and frame its structure in a semi-structured format with the following components:

\begin{enumerate}
  \item \emph{Action Entity} (A): An action entity represents the central objective of the whole argument and is directly derived from the claim as a verbal phrase.
  This way of framing an action entity from claim is motivated by the conclusion of Argument from Consequences scheme which states that ``Action should/shouldn't be bought about''.
%   The action entity is derived directly from claim as a verbal phrase.
  For example, as shown in Figure~\ref{fig:intro_eg}, for the claim ``We should ban surrogacy'', the action can be framed as \emph{``Banning surrogacy''}.
  
  \item \emph{Outcome Entity} (O): An outcome entity represents the consequence of doing an action, where the consequence is either caused or suppressed by the action. 
  The outcome entity is directly derived from the premise with slight modifications in its phrasing.
  For example, as shown in Figure~\ref{fig:intro_eg}, for the premise ``Surrogacy often creates abusive and coercive conditions for women'', the outcome can be framed as \emph{``Abusive and coercive conditions for women''} such that it forms the following relation: \emph{``Banning surrogacy''} $\xrightarrow{\text{suppress}}$ \emph{``Abusive and coercive conditions for women''.}

  \item \emph{Implicit Causal Knowledge} (I):
  In order to understand why/how premise offers support to the claim, we need to explicate knowledge that is either missing or implicit in the argument.
%   we need knowledge, i.e., knowledge that is either missing or is implicit.
  Specifically, we need knowledge that explains the causal connection between action and outcome entities such that the reasoning link between claim and premise becomes clear.
  For example, the implicit knowledge i.e., \emph{``decrease in number of women working as surrogates''} (as shown in Figure~\ref{fig:intro_eg}) is required to understand why/how banning surrogacy suppresses abusive and coercive conditions for women. 
  We term such knowledge as \emph{Implicit Causal Knowledge} and represent it along with the action and outcome entities in the following form:
  
    \begin{itemize}
        \item Banning surrogacy $\xrightarrow{\text{cause}}$ \emph{Decrease in number of women working as surrogates.}
        \item \emph{Decrease in number of women working as surrogates} $\xrightarrow{\text{suppress}}$ Abusive and coercive conditions for women.
    \end{itemize}
  
  \item \emph{Causal relation}: The causality between action entity, outcome entity and implicit causal knowledge is represented with \emph{cause/suppress} labels. 
  Although, the expressibility of the implicit reasoning will be reduced by employing pre-defined causal labels, we hypothesize that majority of typical instances of implicit reasoning in arguments can be captured by encoding such causal labels.
%   Although it's important to consider other types of causality as much as possible, we assume that cause and suppress labels offer a more appropriate degree of expressiveness as well as restriction in framing the implicit reasoning.
\end{enumerate}

Figure~\ref{fig:intro_eg} shows how the final implicit reasoning can be represented in a semi-structured format along with the other aforementioned components.
The template for framing such implicit reasoning can be depicted as \emph{A}$\xrightarrow{\text{cause/suppress}}$ \emph{I}$\xrightarrow{\text{cause/suppress}}$ \emph{O}, where the following relationship is expected:
\begin{itemize}
    \item If $ A \xrightarrow{\text{cause}} I$ and $ I \xrightarrow{\text{cause}} O $, then $A \xrightarrow{\text{cause}} O$
    \item If $ A \xrightarrow{\text{cause}} I$ and $ I \xrightarrow{\text{suppress}} O $, then $A \xrightarrow{\text{suppress}} O$
    \item If $ A \xrightarrow{\text{suppress}} I$ and $ I \xrightarrow{\text{cause}} O $, then $A \xrightarrow{\text{suppress}} O$
    \item If $ A \xrightarrow{\text{suppress}} I$ and $ I \xrightarrow{\text{suppress}} O $, then $A \xrightarrow{\text{cause}} O$
\end{itemize}

\section{Task Design and Annotation}

We design a two phase annotation scheme to obtain semi-structured implicit reasoning at a large-scale, where each phase (\S~\ref{phase_1} and \S~\ref{phase_2}) can be operationalized with crowdsourcing.
In addition, we implement several mechanisms, e.g., sanity checks throughout the task for quality assurance and to ensure that annotators perform the task as instructed.

\subsection{Phase 1: Framing semi-structured implicit reasoning}
\label{phase_1}
In order to frame semi-structured implicit reasoning, we need four main components (\S~\ref{sec-3}) i.e., \emph{Action entity}, \emph{Outcome entity}, \emph{Implicit causal knowledge} and \emph{causal relations}. 
Since an action entity can be derived from its claim uniformly, we automatically derive it as a verbal phrase through a simple rule based pattern matching via Spacy~\cite{spacy}. 
For example, the action entity \emph{``Introducing compulsory voting''} can be derived from the claim \emph{``We should introduce compulsory voting''}.
To collect the remaining components, we design a crowdsourcing task that contains two consecutive steps, namely (1) Annotating outcome entity followed by (2) Annotating implicit causal knowledge, which are described as follows.

\begin{figure}[!htb]
    \centering
    % \vspace*{+0.5cm}
    \includegraphics[width=\textwidth,height=0.88\textheight]{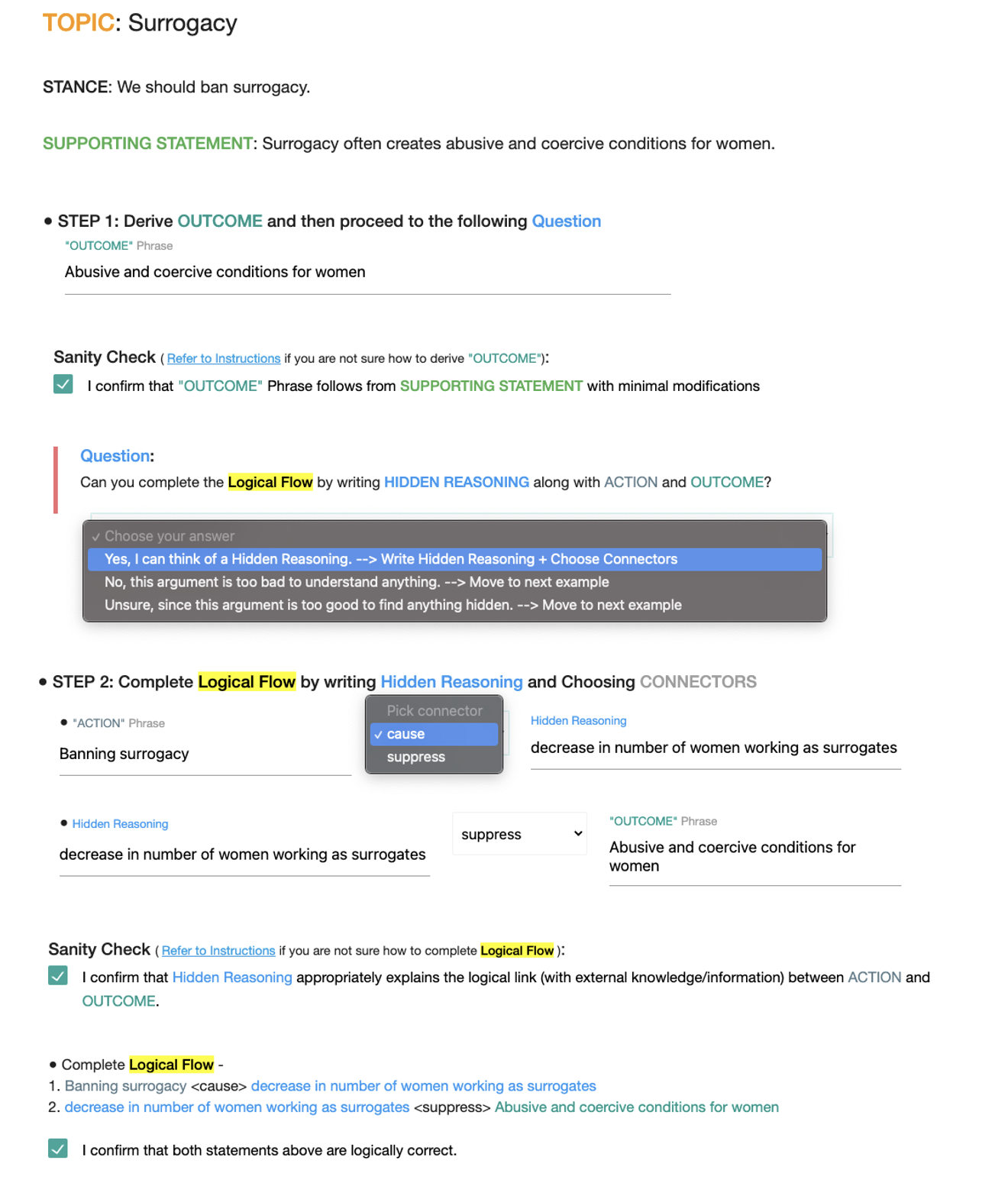}
    \caption{The interface of our crowdsourcing task for phase 1. This phase consists of two steps, where STEP 1 is mandatory while STEP 2 depends on the choice made by crowdworkers for the Question preceding STEP 2.}
    \label{fig:interface}
\end{figure}

\paragraph{Annotating outcome entity} For this step, as opposed to deriving the outcome entity automatically, we leverage crowdsourcing.
We assume that there can be multiple ways one can perceive the consequence of doing an action (i.e., consequence caused or suppressed by action) which can assist in capturing diverse implicit reasoning.
For example, given a Claim: \emph{We should abolish intellectual property rights} and Premise: \emph{People or companies owning the rights to certain ideas can create a closed market, where the owners of such ideas are able to set the price without the fear of competition.}, there can be more than one way to derive the outcome entity and annotate the relation between action and outcome entity: (i) \emph{``Abolishing intellectual property rights''} $\xrightarrow{\text{suppress}}$ \emph{``Creation of a closed market''.} and (ii) \emph{``Abolishing intellectual property rights''} $\xrightarrow{\text{cause}}$ \emph{``Fear of competition''}, which may consequently result in differently framed implicit reasoning.

An example annotation is shown in Figure~\ref{fig:interface}, where in STEP 1 crowdworkers are asked to derive the outcome entity for a given premise~\footnote{We avoid using complicated jargon in our crowdsourcing interface in order to make the task easier for crowdworkers to understand. 
Specifically, we refer to Claim as Stance, Premise as Supporting statement, Implicit causal knowledge as Hidden reasoning and Causal relations as Connectors}.
We additionally instruct crowdworkers to make sure that the derived outcome entity (i) conveys the same meaning as stated in the premise and (ii) represents the consequence of doing an action, for example, \emph{"Banning surrogacy"}$\xrightarrow{\text{suppress}}$ \emph{``Abusive and coercive conditions for women''.} so as to avoid noisy annotations.

\paragraph{Annotating implicit causal knowledge}
The task of annotating implicit causal knowledge is shown as STEP 2 in Figure~\ref{fig:interface}.
For this step we assume that annotation of such knowledge may not be possible for every given claim and premise pair. 
Specifically, for a bad premise there may be no feasible way to explicate the implicit reasoning connection between claim and premise.
For example, given a Claim: \emph{``We should introduce multi-party system''} and Premise:  \emph{``Introducing multi-party system is the right thing to do''}, it is not possible to write any implicit causal knowledge since the argument is a fallacy (i.e.  begging the question) where premise provides no adequate support to the claim.
Similarly, for arguments with very good premise, it may not be necessary to annotate any implicit causal knowledge since it might already be explicated in the premise.
In order to handle such instances, prior to this step, we explicitly ask annotators to judge the feasibility of framing the implicit causal knowledge with given action entity and their annotated outcome entity (See ``Question'' in Figure~\ref{fig:interface}).
This is rather tricky since annotators may be biased to answer "No" or "Unsure" to avoid doing the task and complete the task quickly. 
To avoid such case and reduce bias, we treat this as a bonus question and grant bonus depending on majority response i.e., if majority of crowdworkers believe that an implicit causal knowledge can be explicated or if majority of crowdworkers believe otherwise.

An example annotation for STEP 2 is shown in Figure~\ref{fig:interface}, where crowdworkers are provided with pre-defined templates for framing the relationship between action entity, outcome entity and implicit causal knowledge along with causal relations.
Instead of framing the template as a single chain, we rephrase it into individual relations as: (i) \emph{Action Entity}$\xrightarrow{\text{cause/suppress}}$ \emph{Implicit Causal Knowledge} and (ii) \emph{Implicit Causal Knowledge} $\xrightarrow{\text{cause/suppress}}$ \emph{Outcome Entity}, where the final implicit reasoning can be rephrased as  \emph{Action Entity}$\xrightarrow{\text{cause/suppress}}$ \emph{Implicit Causal Knowledge} $\xrightarrow{\text{cause/suppress}}$ \emph{Outcome Entity}.

\paragraph{Annotating causal relations}
As shown in Figure~\ref{fig:interface}, the annotation of causal relations between components is done alongside the annotation of implicit causal knowledge. 
Crowdworkers are asked to pick one out of two choices of causal relations (i.e., cause and suppress) to form the causal connection between \emph{(Action entity and Implicit causal knowledge)} and \emph{(Implicit causal knowledge and outcome entity)}. 
We include additional sanity checks (See Figure~\ref{fig:interface}) with the final annotated implicit reasoning for crowdworkers to confirm their annotation.

\subsection{Phase 2: Qualitative filtering}
\label{phase_2}

\begin{table}[t]
    % \small
    \begin{tabularx}{\columnwidth}{l|X}
        \textbf{Score} & \textbf{Explanation} \\
        \toprule
         \textbf{1} & Hidden reasoning is completely nonsensical and fails to explain the reasoning link between \emph{Action} and \emph{Outcome}. \newline
         OR \newline
         The use of both connectors is logically incorrect. \\
         \midrule
         \textbf{2} & Hidden reasoning is related to the argument but is a paraphrase of the Stance/Supporting Statement.\newline
         OR \newline
         The use of one or more connectors is logically incorrect. \\
         \midrule
         \textbf{3} & Hidden reasoning is related to the argument but instead of explaining the reasoning link between \emph{Action} and \emph{Outcome}, presents a new supporting statement.\newline
         OR \newline
         The use of one or more connectors is logically incorrect. \\
         \midrule
         \textbf{4} & Hidden reasoning explains the reasoning link between \emph{Action} and \emph{Outcome} fairly good but needs some improvements in wordings. \newline
         AND \newline
         The use of both connectors is logically correct.\\
         \midrule
         \textbf{5} & Hidden reasoning makes it easy to understand the reasoning link between \emph{Action} and \emph{Outcome}. \newline
         AND \newline
         The use of connectors is logically correct.\\
        \bottomrule
    \end{tabularx}
    \caption{Guidelines used by crowdworkers for  phase 2 of our annotation scheme, where they are instructed to score the quality of implicit reasoning on a scale of 1-5.}
    \label{tab:filter}
\end{table}
In order to confirm that the implicit reasonings obtained from phase 1 are indeed of high quality and correct, we perform phase 2 annotation to filter only those implicit reasoning that fulfill the following requirements: (i) Annotated outcome entity actually follows from the premise and (ii) Annotated implicit reasoning is composed of logically correct causal relations along with correctly explicated implicit causal knowledge that makes the reasoning link between action and outcome entities clear. 

At this phase, firstly, we filter the implicit reasonings which do not fulfill requirement (i) i.e., we remove implicit reasonings for which outcome entity does not follow from the premise.
We assume that if the outcome entity does not follow from the given premise, then the implicit reasoning cannot be considered valid. 
For this, five crowdworkers are asked to judge if the outcome entity actually follows from the given premise and are given "Yes/No" choice. 
Then, their judgements are aggregated by majority vote to determine a valid outcome entity. 
Secondly, in order to filter the implicit reasoning which does not meet our requirements (ii), we collect annotations of scores according to the guidelines as shown in Table~\ref{tab:filter}.
Implicit reasonings with valid outcome entities are scored by five crowdworkers on a scale of 1-5 and the final score is decided by majority vote such that implicit reasonings that receive a majority score of 4 or 5 (i.e., three or more than three crowdworkers give a score of 4 or 5) are kept while the rest are discarded.
In order to avoid biased judgements, for phase 2 annotation, we hire additional set of crowdworkers who did not participate in the annotation of implicit reasonings in phase 1.

\section{Crowdsourcing Semi-structured Implicit Reasoning}
% In order to acquire implicit reasonings for arguments, 
We choose Amazon Mechanical Turk (AMT)\footnote{\url{https://www.mturk.com/www.mturk.com}} as our crowdsourcing platform due to its success in previous argumentation mining tasks~\cite{habernal-etal-2018-semeval}.
Prior to conducting the main annotations of implicit reasonings, we conduct multiple annotation studies and pilot runs to finalize our crowdsourcing design. 

\paragraph{Source Data}
\begin{wraptable}{r}{0.55\linewidth}
    \small
    \begin{tabular}{l|c}
    \toprule
    Claim: \emph{We should} & \# of premise \\
    \midrule
    Abandon the use of school uniform & 145\\
    Abolish capital punishment & 176\\
    Abolish zoos & 141 \\
    Ban whaling & 164\\
    Introduce compulsory voting & 116\\
    Legalize cannabis & 210\\
    \midrule
    Total & 952 \\
    \bottomrule
    \end{tabular}
    \caption{Summary of source data used in our task.}
    \label{tab:source_data}
\end{wraptable}
We utilize a well-known argumentation dataset, IBM-Rank-30K corpus~\cite{gretz2019large} for our implicit reasoning annotation. The dataset consists arguments in the form of claim and premise which is suitable for our crowdsourcing task.
% instead of creating a corpus of arguments from scratch. 
It consists around 30K crowd-sourced arguments annotated with stance (i.e., support or against a given claim)~\footnote{In our work we only focus on arguments with support stance.} and point-wise quality.
The arguments were collected with strict length limitations and extensive quality control measures. 
We select a subset of 6 common debatable topics out of a total of 71 topics in the dataset for our implicit reasoning annotation task.
This step is necessary to ensure that crowdworkers are indeed familiar with the topic in order to ascertain the quality of annotations.
We then filter arguments of low point-wise quality (below 0.5) and unclear stance (below 0.6) to make sure that arguments of sufficient quality are used for our annotation task.
After these filtering, 952 arguments were yielded for the 6 topics.
% For our 6 topics, we were left with a total of 952 arguments that passed these filters.

\paragraph{Pilot Crowdsourcing}
We conduct multiple pilot tests on AMT to ensure that our task design is suitable for collecting the implicit reasoning annotations from non-expert crowdworkers, and that the crowdworkers understand and can clearly follow the task instructions.
Additionally, in order to address any ethical issues~\cite{adda2011crowdsourcing} raised by our task, we actively monitor the feedback given by the crowdworkers and communicate with them to resolve any questions/comments raised. Crowdworkers are paid in accordance with the minimum wage (\$0.40) during the pilot tests, calculated by conducting many trials and based on their average work-time to ensure fair pay.

\paragraph{Main Crowdsourcing}
Based on our findings from the pilot tests, we only allow annotators who have $\geq$ 98\% acceptance rate and $\geq$ 5,000 approved Human Intelligence Tasks (HITs) for our main annotation tasks (i.e., Phase-1 and Phase-2).
Prior to each main task, we additionally hold a preliminary qualification quiz that consists of several basic questions for testing crowdworkers reasoning skills. 
Workers who score more than a pre-defined threshold ($\geq$ 75\%) are granted access to do our tasks.
Additionally, to encourage diversity in annotations of implicit reasonings, we recruit new crowdworkers throughout the annotation process. 
In total, 37 workers who cleared the qualification quiz were selected for framing implicit reasoning (Phase~1) and 163 workers were selected for qualitative filtering (Phase~2). 
Our qualifications were open to annotators from location in one of the English-speaking countries. The total costs for our crowdsourcing tasks were about \$4,000 including bonuses and trial-runs.

\section{Preliminary dataset of implicit reasonings}
\begin{table}[t]
    \small
    \begin{tabular}{p{0.6\textwidth}|p{0.15\textwidth}|p{0.15\textwidth}}
    & Phase~1 & Phase~2 \\
    \toprule
    \# of implicit reasonings & 932 & 443 \\
    \# of unique implicit reasonings & 831 & 398\\
    \% of premise with implicit reasonings & 90\% & 79.60\% \\
    Avg. \# of implicit reasonings/premise & 4.14(225)  & 2.23(199)\\
    \midrule
    \# of premise with no implicit reasoning & 25 & 51\\
    \# of premise with one implicit reasoning & 0 & 63\\
    \# of premise with multiple implicit reasoning & 225 & 136\\
    \midrule
    Avg. outcome entity length (words) & 5.47 & 5.75\\
    Avg. premise length (words) & 17.61 & 17.78\\
    Avg. implicit reasoning length (words) & 5.56 & 5.70\\
    \bottomrule
    \end{tabular}
    \caption{Statistics of our preliminary dataset of implicit reasonings at the end of each crowdsourcing phase.}
    \label{tab:statistics}
\end{table}
We perform crowdsourcing annotation on a set of 250 arguments\footnote{Since this is a work-in-progress, we plan to collect implicit reasonings for the total 952 arguments eventually.} covering 6 topics.
For each argument, we allow a maximum of 5 annotators in both Phase~1 and Phase~2.
The cost of annotation per worker is \$0.50 without bonus and \$0.75 with bonus in Phase~1 whereas in Phase~2, the cost of annotation per worker is \$0.40.

\subsection{Dataset Statistics}

We present dataset statistics at the end of each annotation phase in Table~\ref{tab:statistics}. 
At Phase~1, for each argument, it is possible to collect at most 5 implicit reasonings (given that every crowdworker writes an implicit reasoning).
However, not every crowdworkers chose to write an implicit reasoning. 
For each claim-premise pair, after aggregating the majority response i.e. if more than 3 out of 5 annotators agree that they can complete an implicit reasoning, we keep such annotations and discard the implicit reasonings otherwise. 
We additionally discard the implicit reasonings for which crowdworker's response is indecisive. 
Overall, we find that in Phase~1, out of 250 claim and premise pairs, the annotators had a majority response for 225 claim-premise pairs resulting in 932 annotated implicit reasonings.
% Out of the remaining 25, 9 premise were marked as too good to write an implicit reasoning and 2 were marked as bad quality premise for which it's not feasible to write any implicit reasoning. 
% The remaining 14 were marked as indecisive due to no clear majority.

We use the 932 implicit reasoning for Phase-2 qualitative filtering.
After filtering the implicit reasonings which do not meet our requirements~(See~\S~\ref{phase_2}), our count of implicit reasonings reduces down to 443 indicating that Phase~2 might be necessary to filter out the bad implicit reasonings.
% Moreover, we find that the annotated implicit reasonings are short (<5 words) indicating the language used is simple.
% Similarly, the outcome entities are short (<5 words) as compared to premise (<17 words) indicating that outcome contains only key information from premise.
Specifically in Phase-2, first we filter out implicit reasonings that have outcome which actually does not follow from the given premise.
For example, the outcome "Proper judgement" does not follow from the premise "cannabis can help cure people with several problems, legalizing it will provide people with a better quality product". 
We filter such annotation by a majority voting i.e. if more than 3 workers say that the outcome is bad, we remove it along with the annotated implicit reasoning. 
Overall, we find that 103 implicit reasonings had incorrect outcome while the remaining 829 had correctly derived outcome.
Next, we aggregate the scores (1-5) given to the remaining 829 implicit reasoning. 
We find that out of 829 implicit reasoning, 294 are given a majority score of 3 or less and the remaining 443 are given a majority score greater than 3, indicating good quality implicit reasoning.
Additionally, we find 92 implicit reasonings as doubtful i.e., no majority score can be derived and therefore, we discard them. 

\subsection{Data Quality}
\begin{wrapfigure}{r}{0.5\linewidth}
    \centering
    \includegraphics[width=0.8\linewidth]{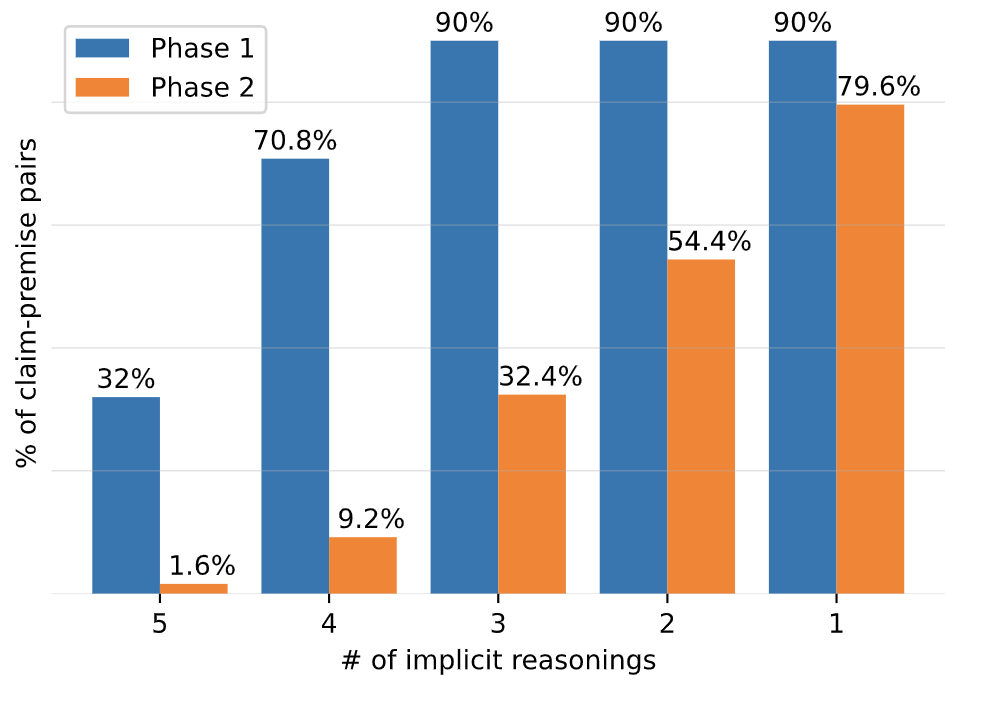}
  \caption{Claim-premise coverage per number of implicit reasonings}
  \label{fig:coverage}
\end{wrapfigure}

\paragraph{Coverage of Implicit Reasonings}
We assess the coverage of our implicit reasonings at the end of Phase 1 and Phase~2 as shown in Figure~\ref{fig:coverage}.
Specifically, we would like to figure out if our causality based annotation template can be used to explicating implicit reasoning for a wide range of arguments.
We find that at the end of Phase-2, there is at least one implicit reasoning for 79.6\% claim-premise pairs while almost half of the arguments have two implicit reasonings annotated.
% We also observe that as the number of implicit reasonings possible for a claim-premise pair increases, the coverage, i.e., the number of arguments having that many implicit reasoning decreases.
Although at Phase~1, the coverage of implicit reasoning is high, it might be due to the inclusion of bad quality implicit reasonings.
In total, we find that for 199 premise (i.e. 79.6\% of all claim-premise pairs) at least one implicit reasoning is annotated with an average of 2 annotations per claim-premise pair as shown in Table~\ref{tab:statistics}.

\paragraph{Quality of Implicit Reasonings}
We randomly sampled 100 implicit reasoning obtained after qualitative filtering of Phase~2 and asked three experts to repeat the same process as explained in Phase~2. 
% Experts judged if the implicit reasoning is of high quality with correct usage of causal labels and appropriate implicit causal knowledge. 
After aggregating experts annotation, we obtain an Krippendorff's $\alpha$~\cite{krippendorff2011computing} of 0.46 for the task of identifying if premise follows from outcome and $\alpha$ (interval metric) of 0.44 for scoring the implicit reasoning on a scale of 1-5 which depicts moderate to substantial agreement.
After further analysis, we find that experts judged 45 implicit reasoning annotations out of 100 to be of high quality.
% Additionally, we also ask experts to judge if the outcome entity can be derived from the premise, but found no cases where outcome entity cannot be derived from its premise indicating that crowdworkers annotations can be considered reliable.

% \input{Table/final_example.tex}
\section{Conclusion and Future Work}
In this work, we present a novel annotation scheme to crowdsource semi-structured implicit reasonings which explain the connection between claim and premise. 
Overall we find high coverage of our annotated implicit reasoning although after our expert evaluation we observe that more than half of the annotated implicit reasoning might be of low quality.
We attribute this difference to be due to either our majority based filtering approach or workers are not performing the task as required. 
Since this is an ongoing work, in future, we will try to employ more advanced techniques (e.g. MACE~\cite{hovy-etal-2013-learning}) to aggregate and filter high quality implicit reasonings.

% \begin{table}[t]
% %    \small
%     \begin{tabular}{l|c|c}
%     \toprule
%     Expert(E1, E2, E3) & Step 1 & Step 2 \\
%     \midrule
%     E1, E2 & 0.52 & 0.50 \\
%     E1, E3 & 0.39 & 0.31 \\
%     E2,E3 & 0.47 & 0.50 \\
%     \midrule
%     E1, E2, E3 & 0.46 & 0.44 \\
%     \bottomrule
%     \end{tabular}
%     \caption{Krippendorff's Alpha between expert annotators for judging whether annotated outcome comes from premise (Step 1) and scoring the implicit reasoning on scale of 1-5 (Step 2).}
%     \label{tab:source_data}
% \end{table}

% \begin{table}[t]
% %    \small
%     \begin{tabular}{l|c|c|c}
%     \toprule
%      & E1 & E2 & E3 \\
%     \midrule
%     \# of instances with correct outcome (Step 1) & 85 & 73 & 65 \\
%     \# of instances with score >= 4 (Step 2) & 66 & 46 & 46 \\
%     \bottomrule
%     \end{tabular}
%     \caption{Breakdown of experts annotation for 100 randomly sampled gold instances.}
%     \label{tab:source_data}
% \end{table}

\bibliography{sample.bib}
\bibliographystyle{plainnat}

\end{document}